\documentclass[10pt,twocolumn,letterpaper]{article}

\usepackage{wacv}              % To produce the CAMERA-READY version
\usepackage{graphicx}
\usepackage{amsmath}
\usepackage{amssymb}
\usepackage{booktabs}

\usepackage[pagebackref,breaklinks,colorlinks]{hyperref}

\usepackage[capitalize]{cleveref}
\crefname{section}{Sec.}{Secs.}
\Crefname{section}{Section}{Sections}
\Crefname{table}{Table}{Tables}
\crefname{table}{Tab.}{Tabs.}

\let\titleold\title
\renewcommand{\title}[1]{\titleold{#1}\newcommand{\thetitle}{#1}}
\def\maketitlesupplementary
   {
   \newpage
       \twocolumn[
        \centering
        \Large
        \textbf{\thetitle}\\
        \vspace{0.5em}Supplementary Material \\
        \vspace{1.0em}
       ] %< twocolumn
   }

\def\wacvPaperID{2240} % *** Enter the WACV Paper ID here
\def\confName{WACV}
\def\confYear{2025}

\begin{document}

\title{Disharmony: Forensics using Reverse Lighting Harmonization}

\author{Philip Wootaek Shin$^{1,2}$, Jack Sampson$^1$, Vijaykrishnan Narayanan$^1$, \\ Andres Marquez$^2$, Mahantesh Halappanavar$^2$  \\
$^1$The Pennsylvania State University \\$^2$Pacific Northwest National Laboratory  \\
{\tt\small \{pws5345,jms1257,vxn9\}@psu.edu}, 
 {\tt\small \{andres.marquez,hala\}@pnnl.gov}
% \and
}
\maketitle

%Applications papers will be evaluated on systems-level innovation, novelty of the domain and comparative assessment
%%%%%%%%% ABSTRACT
\begin{abstract}
  Content generation and manipulation approaches based on deep learning methods have seen significant advancements, leading to an increased need for techniques to detect whether an image has been generated or edited. Another area of research focuses on the insertion and harmonization of objects within images. In this study, we explore the potential of using harmonization data in conjunction with a segmentation model to enhance the detection of edited image regions. These edits can be either manually crafted or generated using deep learning methods. Our findings demonstrate that this approach can effectively identify such edits. Existing forensic models often overlook the detection of harmonized objects in relation to the background, but our proposed \textbf{Disharmony} Network addresses this gap. By utilizing an aggregated dataset of harmonization techniques, our model outperforms existing forensic networks in identifying harmonized objects integrated into their backgrounds, and shows potential for detecting various forms of edits, including virtual try-on tasks.
  
\end{abstract}

% \baselineskip

%%%%%%%%% BODY TEXT
\section{Introduction}
\label{sec:intro}

The rapid advancements in content generation and manipulation approaches based on deep learning methods (\textit{Generative AI}, broadly) have led to a proliferation of tools capable of creating highly realistic images\cite{DallE3,StableDiffusion} or editing existing images\cite{Instructpix2pix, Imagic}. As these technologies become more sophisticated, the ability to distinguish between real and AI-generated or edited images becomes increasingly challenging. This has sparked a growing interest in developing methods to detect such alterations, which is crucial for applications in digital forensics, media authenticity, and content verification. A key class of image manipulations of interest to forensics is the seamless insertion and harmonization of objects into images, and this has been a notable target for deep learning models. Harmonization ensures that the inserted object blends naturally with the surrounding environment, making it difficult to distinguish from the original content. While significant progress has been made in generating harmonized images, there remains a gap in the detection of these subtle edits, especially when traditional forensic models are employed.

\begin{figure}[!t]
    \begin{center}
        \includegraphics[width=\linewidth]{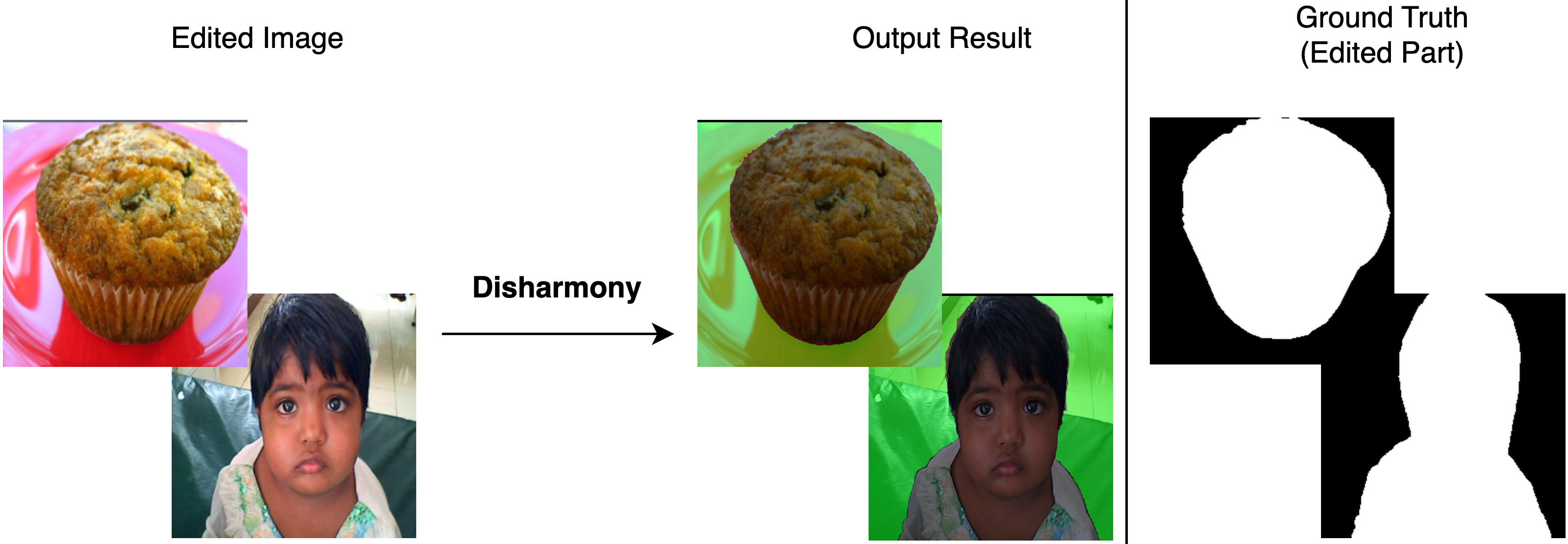}
    \end{center}
\vspace{-12pt}\caption{Task overview of our Disharmony}
\label{fig:task}
\end{figure}

Current forensic detection models~\cite{IML-ViT,HiFi-net,CAT_net,Mantra-net} often struggle to identify harmonized objects, particularly when these objects are skillfully integrated into the background. However, performing lighting harmonization has been a historically difficult challenge, and even modern deep-learning-based approaches that are increasingly effective at fooling the naked eye could well be leaving artifacts in their wake. Such artifacts have been discovered for other editing approaches~\cite{syntheticImageVerification}. We propose the \textit{Disharmony} network, a Harmonization Detection Model specifically designed to detect various forms of image edits. Fig.~\ref{fig:task} illustrates our Disharmony Network's approach to the new task of detecting edited parts of an image based on light discrepancies. Specifically, we explore whether training models on harmonization tasks, coupled with segmentation models, can enhance their ability to identify areas where objects have been added or modified. Our study focuses on detecting edits such as the insertion of new objects followed by harmonization, manual photo manipulation using tools like Photoshop\cite{adobephotoshop}, as well as edits generated by deep learning models, including diffusion models. Our research demonstrates the effectiveness of this approach, offering a novel solution to the challenges posed by the increasingly sophisticated capabilities of generative AI. 
%This paper investigates the potential of using harmonization data, coupled with segmentation models, to detect edited regions within images. 
Our contributions are as follows: 
\begin{itemize}
\item We evaluated existing harmonization methods that involve object insertion followed by harmonization, testing these with current forensic methods to assess their detectability. Our results support the need for forensic models more capable of discerning harmonized edits.
\vspace{-6pt}\item We trained a segmentation model using an aggregated dataset consisting of three distinct harmonization methods: light adjustment via a neural network-based approach, light adjustment via a physics-based method, and light adjustment through handcrafted, visually appealing techniques. Our results demonstrate that the aggregated approach significantly outperforms existing forensic methods on harmonized data.
\vspace{-6pt}\item  We conducted a sensitivity study on our network using a test set incorporating various harmonization methods and showed that Disharmony can be extended for use beyond edits performed by the three harmonization approaches used for training.

\end{itemize}

%------------------------------------------------------------------------

\section{Related Work}
\label{sec:Related Work}
% Understanding the context and foundation of our research requires an exploration of two key areas: image harmonization and image forensics. Image harmonization refers to the process of blending inserted objects seamlessly into a target image, ensuring that the added elements appear natural and consistent with the surrounding environment. This technique is crucial for various applications in computer graphics and generative AI, where realism is paramount.

% On the other hand, image forensics involves the development of methods to detect manipulations and alterations within digital images. As the ability to edit images becomes more sophisticated, the need for robust forensic techniques has grown, especially in identifying subtle modifications like those introduced through harmonization. In the following sections, we delve deeper into these two areas, examining the current state of research and their relevance to our study.

To fully grasp the foundation of our research, it is essential to explore three critical areas: image harmonization, image forensics, and image edits via diffusion models. Image harmonization involves seamlessly blending inserted objects into a target image, ensuring that these additions appear natural and consistent with their surroundings—a process crucial for applications requiring high levels of realism in computer graphics and generative AI.

Image forensics, on the other hand, focuses on developing methods to detect manipulations and alterations in digital images. As editing techniques become increasingly sophisticated, the demand for robust forensic methods to identify subtle modifications, including those introduced through harmonization, has grown.

Lastly, diffusion models have emerged as powerful tools for generating and editing images, offering capabilities such as virtual try-on, text-based image editing, and drag-based manipulation. These advanced techniques pose new challenges for forensic detection, making it necessary to understand their impact on image authenticity.

In the following sections, we will delve deeper into each of these areas, examining their current state of research and their significance to our study.

%-------------------------------------------------------------------------
\subsection{Image Harmonization techniques}

The Harmonizer\cite{Harmonizer} approaches image harmonization by predicting filter arguments through a neural network, allowing it to adjust basic image properties like brightness and contrast to achieve realistic results. DoveNet\cite{DoveNet2020}is a deep image harmonization method that uses a novel domain verification discriminator to align the foreground with the background, ensuring consistency in composite images. It was developed alongside the iHarmony4\cite{iharmony4} dataset, which was created to address the lack of high-quality data for image harmonization, and has been shown to be effective through extensive testing on this dataset. The Harmonization Transformer\cite{IHT} leverages the Transformer's ability to model long-range context dependencies to adjust the foreground lighting in composite images, ensuring compatibility with the background while preserving structure and semantics. HI-Net\cite{INR} is a novel image harmonization method that leverages implicit neural networks to achieve high-resolution, dense pixel-to-pixel harmonization without relying on hand-crafted filters. By decoupling the neural network into parts that capture content and environment, and introducing a Low-Resolution Image Prior network, HINet effectively addresses boundary inconsistencies. PCT-Net\cite{PCTNET} is a versatile image harmonization method that applies pixel-wise color transforms (PCTs) to full-resolution images by predicting parameters using downsampled inputs. We have used these five harmonization methods to create a dataset for testing forensic detection capabilities.

The RealHM dataset\cite{RealHMDataset} is a real-world image harmonization dataset created by expert users to evaluate and benchmark harmonization methods. Careaga \etal \cite{IntrinsicHarm}. introduced a self-supervised illumination harmonization approach that operates in the intrinsic image domain, addressing lighting inconsistencies between the foreground and background in composite images. To validate its effectiveness, a user study was conducted using 50 challenging composites created from Unsplash images, demonstrating that this method achieves enhanced realism compared to state-of-the-art harmonization techniques. The DISK25k dataset\cite{Disk25kdata} is generated using OPA Dataset\cite{OPADataset} by combining foreground objects and background images using a structured process that includes deep image matting with the MatteFormer\cite{Matteformer} and image harmonization using the Harmonizer\cite{Harmonizer}. This approach involves creating segmentation masks, applying morphological operations, and utilizing alpha blending to produce realistic, labeled spliced images that closely mimic real-life manipulations. We have used these three datasets—DISK25k, RealHM, and Careaga \etal(IH Dataset) —to train our model.

% \begin{itemize}
% \item Compared network: Harmonizer\cite{Harmonizer}, DoveNet\cite{DoveNet2020},Harmonization Transformer(IHT)\cite{IHT}, HINet(INR)\cite{INR}, PCT-Net\cite{PCTNET}
% \item Dataset: RealHM dataset\cite{RealHMDataset}, Intrisic harmonization dataset\cite{IntrinsicHarm}, Disk25k\cite{Disk25kdata}, iharmony4\cite{iharmony4}(say not applicable to train)

% \end{itemize}

%-------------------------------------------------------------------------
\subsection{Image Forensics}

MantraNet\cite{Mantra-net} is a unified DNN designed for detecting and localizing various types of image forgeries, including splicing, copy-move, and enhancements, without the need for extra preprocessing or postprocessing. CAT-Net\cite{CAT_net} is a convolutional neural network that detects and localizes image manipulations by analyzing JPEG compression artifacts and discrete cosine transform (DCT) coefficients, which retain crucial information from image acquisition and editing processes. HIFI-Net\cite{HiFi-net} addresses the challenge of unified image forgery detection and localization by introducing a hierarchical fine-grained approach that represents and classifies forgery attributes at multiple levels. This method, which includes a multi-branch feature extractor and specialized localization and classification modules, effectively captures the hierarchical nature of forgery attributes, significantly improving detection and localization accuracy across various benchmarks. IML-ViT\cite{IML-ViT} is a novel Transformer-based approach for Image Manipulation Localization (IML) that leverages high-resolution capacity, multi-scale feature extraction, and manipulation edge supervision to effectively capture non-semantic discrepancies between manipulated and authentic regions. This method, which addresses the limitations of CNNs in long-range and non-semantic modeling, has been shown to outperform state-of-the-art IML methods.

We tested these four forensic methods and Disharmony using five harmonization methods. We were unable to test DiffForensics\cite{Diffforensics} and DiffSeg\cite{Diffseg} due to the lack of released source code, despite their reported effectiveness in recent studies. Additionally, we attempted to run the methods proposed by Corvi \etal
\cite{corvi2023intriguing}. and Uhlenbrock \etal \cite{Palette}, but did not achieve desirable results relevant to our work, further emphasizing the need for specialized approaches in this domain.

% \begin{itemize}
% \item Discuss Synthetic Image Verification of Generative Work\cite{SyntheticImage}, Did you note my Pallet\cite{Palette}, DiffForensics(CVPR2024)\cite{Diffforensics}, DiffSeg\cite{Diffseg}
% \item Compared Network: Mantra-net\cite{Mantra-net}, CAT-net-V2\cite{CAT_net},  HiFi-Net\cite{HiFi-net}, IMLViT\cite{IML-ViT}

% \end{itemize}
% \vspace{-9pt}
\subsection{Different Image Edit Methods}
\label{sec:Different Image Edit}

% \vspace{-9pt}

In the realm of image editing, several well-known techniques are employed. The most straightforward approach involves inserting an object into a background and applying perceptual or manually crafted edits to achieve a harmonious appearance\cite{RealHMDataset}. Another approach leverages recent deep learning models, where an object is placed into an image and then harmonized with the background to create a seamless integration\cite{Harmonizer,DoveNet2020,HiFi-net,IHT,PCTNET}.

Moreover, diffusion models have expanded the possibilities of image editing. For instance, models like Imagic\cite{Imagic} and InstructPix2Pix\cite{Instructpix2pix} enable image modifications through text-based instructions. Additionally, virtual try-on applications, such as StableVINTON\cite{stableviton}, allow users to digitally try on clothes from different images onto a model. Recent trends also include techniques where objects or facial features are edited by dragging, such as DragDiffusion\cite{dragdiffusion}, allowing for adjustments in orientation or expression while maintaining the integrity of the object.

In our work, we focus on the first approach, where objects are inserted and then harmonized with the background. While we have conducted preliminary experiments on text-based image editing, virtual try-on, and drag based edits, these areas will be the focus of our future research.

% \begin{itemize}

% \item image edit via text: imagic\cite{Imagic}, instructpix2pix\cite{Instructpix2pix}
% \item virtual try-on StableVITON\cite{stableviton}
% \textcolor{red}{\item edit drag: dragdiffusion\cite{dragdiffusion} free drag\cite{freedrag}}

% \end{itemize}

\begin{figure*}[t!]
\begin{center}
\includegraphics[width=0.9\linewidth]{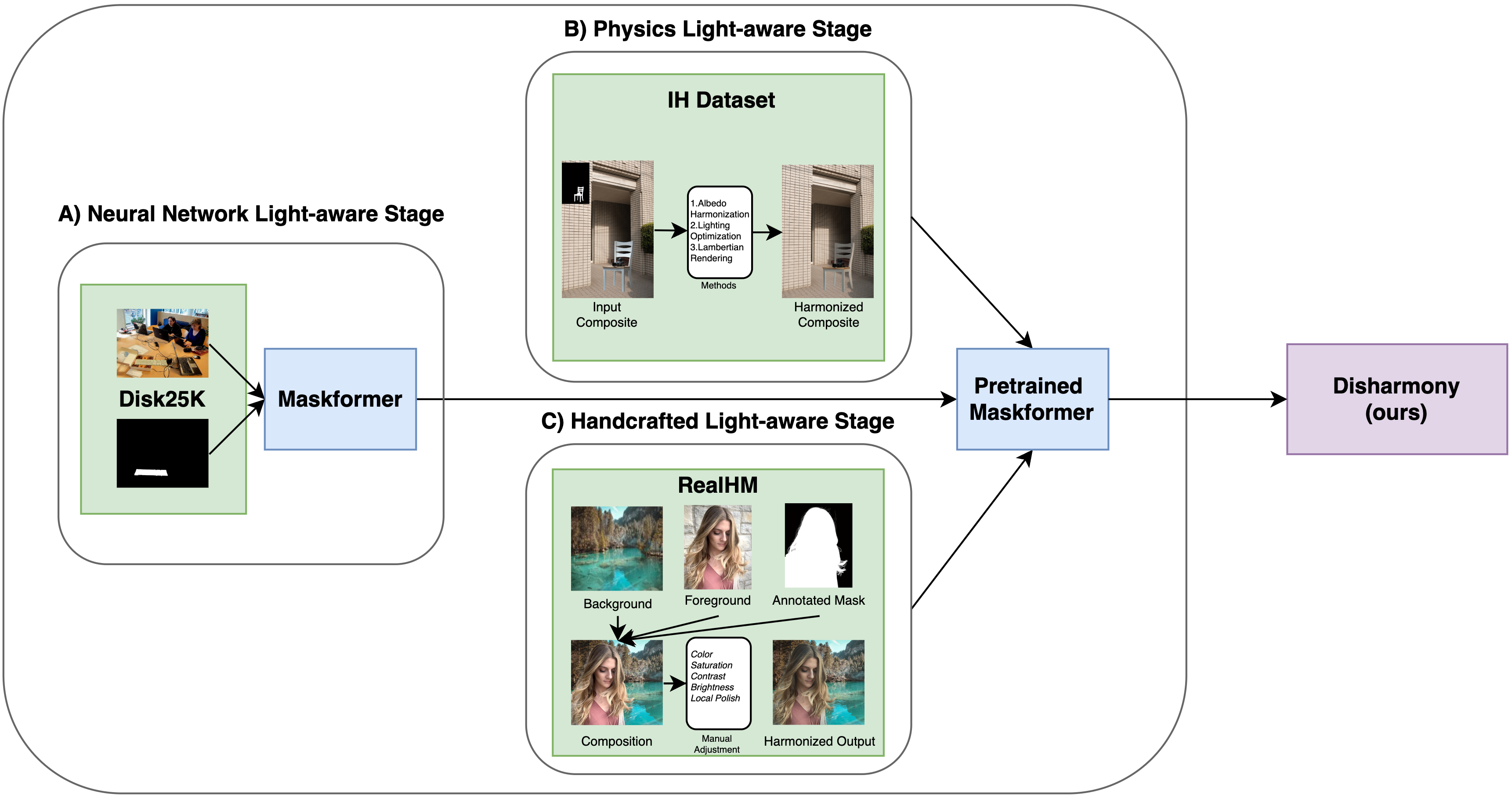}
\end{center}
\vspace{-12pt}
\caption{The overall training pipeline of Disharmony is depicted here. The training process is divided into two stages: initially, we pretrain MaskFormer using the DISK25k dataset(Step A). Subsequently, we fine-tune this model with the IH Dataset(Step B) and the RealHM Dataset(Step C), culminating in the development of Disharmony.}
\label{fig:Training_Pipeline}
\end{figure*}

% \subsection{Test Dataset Generation}
% \label{subsec:TESTDATASET}

%------------------------------------------------------------------------
\section{Method}
\label{sec:Method}

Inspired by DiffSeg\cite{Diffseg}, which utilizes not only raw pixel values but also noise and frequency information to detect and localize edited regions, we hypothesized that even visually appealing edits might leave subtle traces that can be detected. These traces, although small, could potentially be captured and learned by segmentation models such as MaskFormer\cite{MaskformerV1} or SegFormer\cite{xie2021segformer}, enabling effective identification and localization of edited areas. We selected MaskFormer\cite{MaskformerV1} as it proved to be an effective aggregator.

\subsection{Aggregation of Lighting Methods}

% \textcolor{red}{We chose Maskformer\cite{MaskformerV1} as our backbone pretrained model since MaskFormer has following characteristic: The task involves both instance and semantic segmentation (panoptic segmentation). There's a need for a unified model that can handle multiple segmentation tasks. Since we are dealing with multiple objects but only restricting the mask as lighting condition this network was more suitable than the Segformer\cite{xie2021segformer}.}

%\textcolor{red}{ as our backbone pretrained model due to its advantageous characteristics. Rather than relying solely on a per-pixel classification approach, MaskFormer's mask classification formulation offers a more effective way to capture and harmonize inserted objects within an image.} 
% Additionally, MaskFormer allows for the simultaneous handling of multiple objects while restricting the masks to reflect lighting conditions, making it particularly well-suited to our specific requirements}

We identified three distinct methods for editing objects in relation to their background. The first method employs neural networks, specifically using an encoder-decoder structure, to train a model that learns to adapt and harmonize objects with varying backgrounds. The second method involves manual adjustments, where tools like Photoshop\cite{adobephotoshop} are used to modify composites by altering color, saturation, brightness, and other visual parameters to create a result that is aesthetically pleasing to the human eye. The final method is based on traditional and physics-based lighting techniques, which utilize the physical characteristics of light, such as shading, occlusion, and illumination, to adjust the lighting and achieve realistic harmonization.

We hypothesized that aggregating these three methods would enhance the segmentation model's ability to learn lighting features. For the dataset utilizing the encoder-decoder structure, we selected the DISK25k dataset\cite{Disk25kdata}, which comprises 24,964 image pairs of ground truth masks and harmonized images generated using the Harmonizer model. For the manually adjusted dataset, we employed the RealHM dataset\cite{RealHMDataset}, consisting of 216 high-quality, high-resolution foreground/background pairs with corresponding harmonized outputs. These foregrounds include both human portraits and general objects, with backgrounds encompassing diverse environments such as mountains, rivers, buildings, and skies. Additionally, we obtained 60 pairs of composite images, masks, and harmonized images from the dataset used in Careaga(IH Dataset), \etal~\cite{IntrinsicHarm}, which were originally used for testing with human subjects.

\subsection{Training Details}

%For the backbone of our network, we selected MaskFormer. We implemented a two-step training procedure to develop our LightForensics Network, with a focus on leveraging both large-scale learning and domain-specific refinement. 

Fig.~\ref{fig:Training_Pipeline} illustrates Disharmony's training pipeline. Initially, we pretrained our segmentation model using the DISK25k dataset\cite{Disk25kdata}, which contains 24,964 image pairs. This extensive dataset allowed the model to build a robust foundation in understanding basic image harmonization principles, encompassing a wide range of lighting conditions, object-background interactions, and general harmonization patterns. Pretraining on this large dataset enabled the model to acquire a comprehensive skill set in harmonization strategies across various image scenarios.

Following this, we performed transfer learning with the RealHM and Intrinsic Harmonization datasets to refine the model’s capabilities. The RealHM dataset, featuring high-resolution image pairs that emphasize human-centric harmonization, required the model to adapt its learned techniques to tasks closely aligned with human perception. Fine-tuning on RealHM enhanced the model’s ability to meet human expectations in scenarios where the subjective quality of harmonization is crucial. The Intrinsic Harmonization dataset introduced the model to a physics-based harmonization approach, grounded in the physical properties of light and materials. This allowed the model to refine its ability to apply harmonization techniques that are both perceptually pleasing and physically accurate. Fine-tuning on this dataset equipped the model with the ability to integrate harmonization principles that align with real-world physics, thereby adding an extra layer of realism to its outputs. An empirical study of the quality improvements from this approach is included in the supplemental documentation.

While it is generally acknowledged that more data improves model robustness and performance, especially in segmentation tasks related to lighting, we intentionally chose one dataset for each category to evaluate the viability and applicability of our method in forensic detection. This approach allows us to assess whether our method can effectively detect forensic manipulations across different harmonization techniques.

 \begin{figure*}[t!]
\begin{center}
\includegraphics[width=0.95\linewidth]{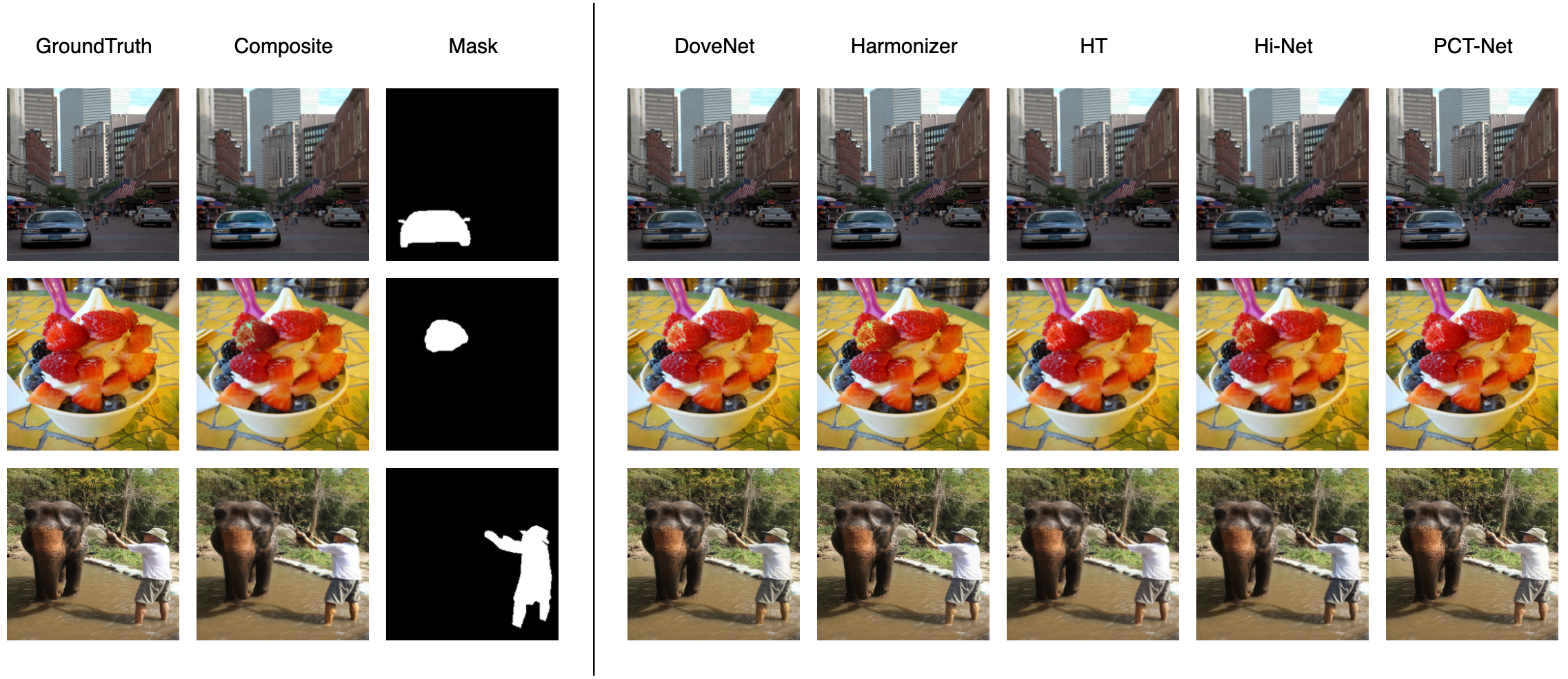}
\end{center}
\vspace{-12pt}
\caption{Samples from the test set. We randomly selected 100 images from the iHarmony4 dataset\cite{iharmony4}, each consisting of a ground truth image, a composite image (randomly chosen one composite from different composites), and a corresponding mask. The composite images and masks were then processed using various harmonization methods (DoveNet\cite{DoveNet2020}, Harmonizer\cite{Harmonizer}, HT\cite{IHT}, Hi-Net\cite{INR}, and PCT-Net\cite{PCTNET}) to generate the test images. }
\label{fig:Test_DATASET}
\end{figure*}

\begin{figure*}[t!]
\begin{center}
\includegraphics[width=0.95\linewidth]{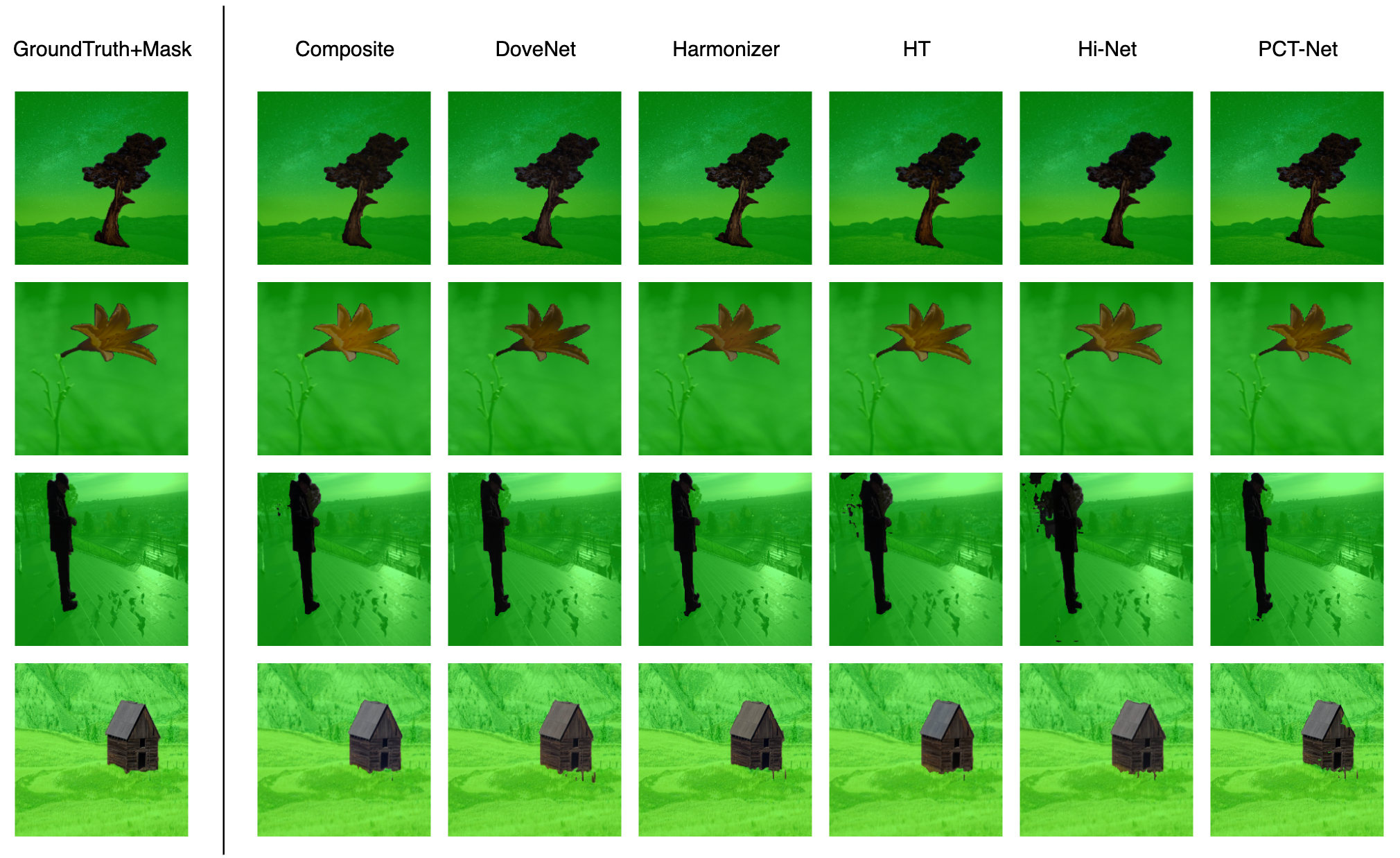}
\end{center}
\vspace{-12pt}
\caption{ Qualitative results of Disharmony across different harmonization methods(DoveNet\cite{DoveNet2020}, Harmonizer\cite{Harmonizer}, HT\cite{IHT}, Hi-Net\cite{INR}, and PCT-Net\cite{PCTNET}) and composite images. For clarity, the background is shown in green to make the segmentation masks more visually discernible, with the resulting segmentation mask retaining the original colors of the image. }
\label{fig:QualEval2}
\end{figure*}

\textbf{Implementation details}: All experiments were conducted on a single NVIDIA GH100 GPU with 100 GB of VRAM. During the first stage of training, the model was trained for 100 epochs, followed by 50 epochs in the second stage.  The model was fine-tuned to receive feature maps of size 512x512 pixels as input, so all images were first resized accordingly before being fed into the network.

\begin{figure*}[p]
\begin{center}
\includegraphics[width=0.91\linewidth]{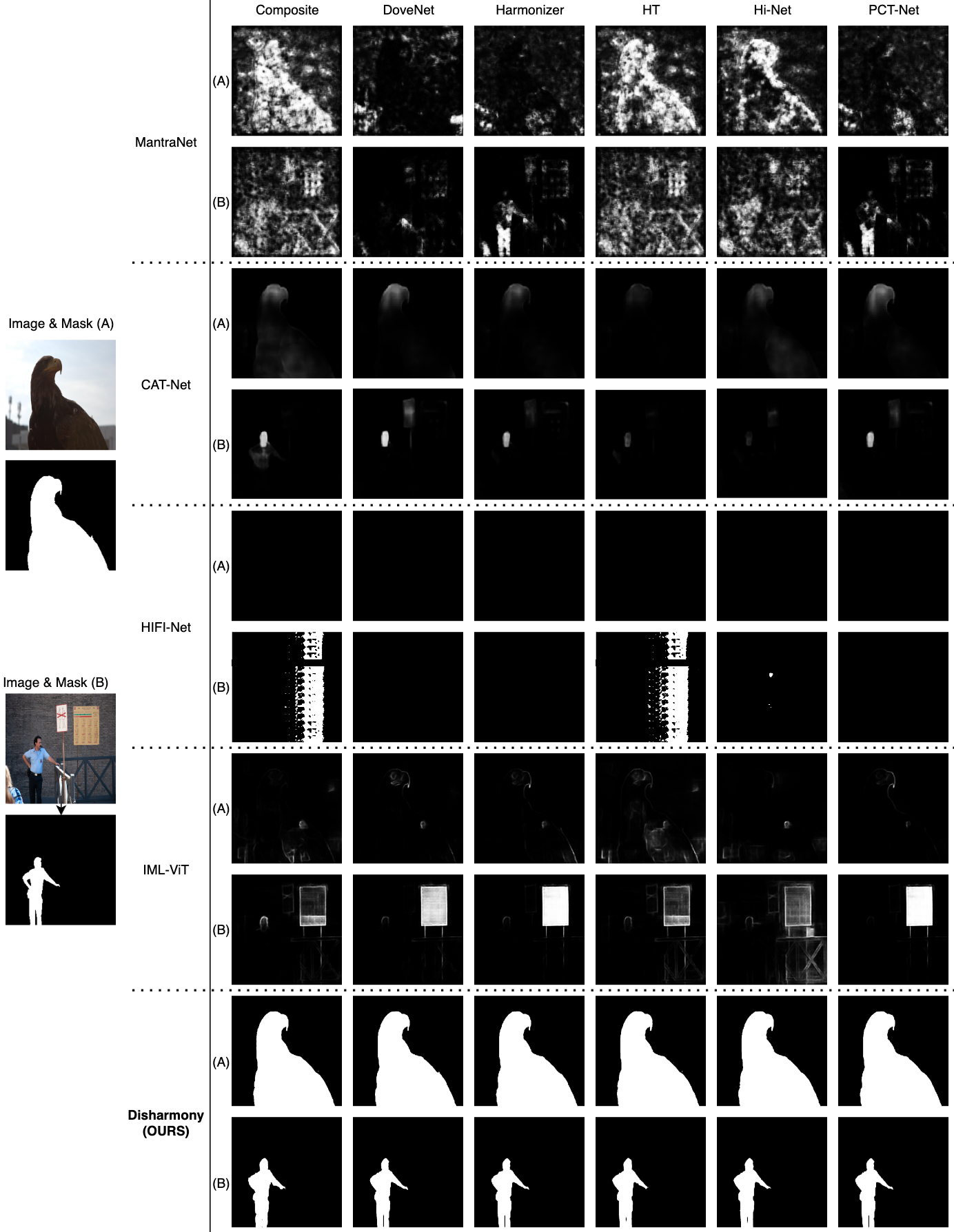}
\end{center}
\caption{Qualitative comparison of various forensic methods(MantraNet\cite{Mantra-net}, Cat-Net\cite{CAT_net}, HiFi-Net\cite{HiFi-net}, IML-ViT\cite{IML-ViT}) and Disharmony across different harmonization methods(DoveNet\cite{DoveNet2020}, Harmonizer\cite{Harmonizer}, HT\cite{IHT}, Hi-Net\cite{INR}, and PCT-Net\cite{PCTNET}), using the given image and mask.  }
\label{fig:QualEval1}
\end{figure*}

\begin{table*}[!ht]
    \centering
    
%   \setlength{\tabcolsep}{3pt}
%   \renewcommand{\arraystretch}{1.1}
    % \small
    \footnotesize
    % \tiny
    % \scriptsize
    % \fontsize{6.2pt}{6.5pt}\selectfont
    \begin{tabular}{l|ccccc}
       \hline
       \textbf{ROC AUC($\uparrow$)} & MantraNet  & CAT-Net & HIFI-NET & IMLViT&\textbf{Disharmony}\\\hline
    Composite & 0.5497 & 0.5595 & 0.5101(4)&0.5723&\textbf{0.7051}\\
    DoveNet  & 0.5394  & 0.5948& 0.4102(3) & 0.5907 &\textbf{0.7144} \\
    Harmonizer  & 0.6774 & 0.5594& 0.4757(4)& 0.6350& \textbf{0.7309} \\
     HT  & 0.5355 & 0.55488 & 0.4353(5)& 0.5511 &\textbf{0.6757} \\
     HI-Net  & 0.5296 & 0.5330& 0.4613(1) & 0.5318 &\textbf{0.6814} \\
    PCT-Net  & 0.5594 & 0.5452&    -              (0) & 0.5661 &\textbf{0.7020} \\
    \hline
    \end{tabular}
    \caption{Evaluation of the image harmonized dataset using different forensic methods (ROC AUC). The x-axis indicates the forensic methods, and the y-axis indicates the harmonization methods. Except for HIFI-NET, all forensic networks reported detections on all 100 inputs. For HIFI-NET, the number of inputs with detections is shown in parentheses.}
    \label{tab:ROCAUC}
\end{table*}

\begin{table*}[!ht]
    \centering
    
%   \setlength{\tabcolsep}{3pt}
%   \renewcommand{\arraystretch}{1.1}
    % \small
    \footnotesize
    % \tiny
    % \scriptsize
    % \fontsize{6.2pt}{6.5pt}\selectfont
    \begin{tabular}{l|ccccc}
       \hline
       \textbf{miou($\uparrow$)} & MantraNet  & CAT-Net & HIFI-NET& IMLViT &\textbf{Disharmony}\\\hline
    Composite & 0.1204& 0.4872 & 0.0037 & 0.0445 &\textbf{0.5801}\\
    DoveNet  & 0.0258 & 0.4992& 0.0041& 0.0554 &\textbf{0.5887} \\
    Harmonizer  & 0.1037 & 0.4714& 0.0004& 0.1063 &\textbf{0.6000} \\
     HT  & 0.1035& 0.4788 & 0.0062& 0.0331&\textbf{0.5543} \\
     Hi-Net  &0.0963 & 0.4659& 0.0& 0.0187&\textbf{0.5638} \\
    PCT-Net  & 0.0210 & 0.4745& 0.0003& 0.0321&\textbf{0.5729} \\
    \hline
    \end{tabular}
     \caption{Evaluation of image harmonized dataset with different forensic works(Miou). The x-axis indicates the forensic methods, and the y-axis indicates the harmonization methods.}
    \label{tab:miou}
\end{table*}

\section{Results and Discussion}

Most image harmonization methods are traditionally trained on the iHarmony4 dataset\cite{iharmony4}, and their performance is often evaluated using qualitative metrics, comparing visual results to demonstrate their effectiveness relative to other methods. However, no existing method has tested harmonized object images against established forensic networks. To address this gap, we created a harmonization object detection test dataset specifically for this purpose.

For testing these forensic models, we extracted a total of 100 images and corresponding masks from the iHarmony4 dataset, which comprises four distinct sub-datasets: HCOCO, HAdobe5K, HFlickr, and HDay2Night. From each sub-dataset, we randomly selected 25 distinct composite images featuring different objects.

After collecting these 100 composite images, we applied five different deep learning-based image harmonization techniques: DoveNet\cite{DoveNet2020}, Harmonizer\cite{Harmonizer}, Harmonization Transformer(HT)\cite{IHT}, Hi-Net\cite{INR}, and PCT-Net\cite{PCTNET}. In addition to the harmonized images, we also included the original composite images from iharmony4 dataset for comprehensive evaluation. The samples from our test set are illustrated in Fig.~\ref{fig:Test_DATASET}.

% \subsection{Comparison of Forensic Methods}

For testing, we selected four different networks: MantraNet\cite{Mantra-net}, CAT-Net\cite{CAT_net}, HIFI-Net\cite{HiFi-net}, and IML-ViT\cite{IML-ViT}, and compared their performance with Disharmony on the test set we generated.

\subsection{Qualitative Evaluation and Comparison}
\label{subsec:Qual_Comp}

We conducted a qualitative evaluation of Disharmony by testing it on five different image harmonization methods, as well as the composite images selected for our test set. The results are illustrated in Fig.~\ref{fig:QualEval2}. As shown, our forensic network is generally effective in detecting the edited regions of images processed by the various harmonization methods. However, there are instances, particularly with Hi-Net and the Harmonization Transformer, where the segmentation results are less sharp. We attribute this to the possibility that our network may be challenged by the specific harmonization techniques employed by these methods, leading to some confusion and less precise segmentation.

Given the results obtained from our network on the test set, we deemed it valuable to compare our performance with that of existing forensic networks. Fig.~\ref{fig:QualEval1} illustrates the comparison. As observed, the existing forensic networks generally exhibit lower forensic distinguishability when applied to images processed by harmonization methods. Notably, HiFi-Net produced anomalous results, with minimal or no detection, despite verifying its performance in other evaluations where it functioned correctly. This highlights the challenges these networks face in effectively handling image harmonization cases.

\subsection{Quantitative Comparison}
\label{subsec:Quant_Comp}

For the quantitative comparison, we utilized the ROC AUC score and mIoU score to evaluate and compare the performance of the forensic networks. The overall performance in terms of AUC scores is presented in Table ~\ref{tab:ROCAUC}, while the mIoU scores are shown in Table ~\ref{tab:miou}.
The mIoU score was chosen as it is a standard metric for evaluating segmentation tasks; however, its main limitation is that the results are dependent on a specific threshold value. We selected a threshold that was empirically determined to produce the most consistent and visually coherent results across the majority of networks, ensuring that the evaluations were fair and reflective of each model's practical performance in detecting and segmenting edited regions. Given that the segmentation task in this study involves only two labels, the AUC score serves as an additional metric that is independent of the threshold value, offering a more comprehensive evaluation of the models' performance.

The results presented in Tables \ref{tab:ROCAUC} and \ref{tab:miou} clearly demonstrate that Disharmony outperformed all other forensic networks across various image harmonization methods. In terms of ROC AUC scores, our network consistently achieved the highest values, significantly surpassing the performance of MantraNet, CAT-Net, HIFI-NET, and IMLViT, with challenging datasets that is generated by different image harmonization methods. Similarly, the mIoU scores further confirm the superiority of our approach, with our network achieving the highest segmentation accuracy across all tested scenarios. While other networks struggled, especially HIFI-NET, which often failed to produce meaningful results, our network demonstrated robust performance, effectively detecting and localizing edited regions in harmonized images. This consistent performance highlights the effectiveness and reliability of our approach in handling diverse image harmonization techniques.

\section{Discussion and Future work}
\label{Dis&Lim}

In certain instances where image edits are limited to a very small portion of the image, none of the evaluated networks, including Disharmony, were robust in accurately detecting these minor changes. Disharmony either failed to identify the image as edited or incorrectly marked other areas as edited. We believe that these limitations could be addressed by fine-tuning the model with datasets that specifically include such edge cases or by incorporating additional data to enhance the model's robustness.

There are various methods for editing images, including the use of diffusion models. These models can generate edited images based on a reference image and a text prompt, with Imagic\cite{Imagic} and InstructPix2Pix\cite{Instructpix2pix}being two prominent examples. We utilized the Imagic TEDBench dataset, which contains 40 images and 100 different edits. For our study, we selected 40 images that did not alter the background relative to the original image. Additionally, we created edited versions of these images using InstructPix2Pix to evaluate whether our forensic network could detect the edits. Table~\ref{tab:comparison_imagicinst} presents the average Structural Similarity Index (SSIM)\cite{SSIM}, and Learned Perceptual Image Patch Similarity (LPIPS)\cite{lpips} scores for the 40 unedited images, compared to their edited counterparts. 

Based on these metrics, we concluded that it would not be fruitful to further explore these edits with Disharmony, as existing techniques are already astute at flagging such heavily altered images and the fundamental nature of the edits diverged from the focus of our forensic detection efforts. We observed that instead of modifying only a specific part of the image, the diffusion models often generated entirely new backgrounds that differed from the original, even though the results appeared visually pleasing. This significantly altered the context of the entire image, rather than simply editing a portion of it, and therefore identifying a specific region of change within the image was frequently an ill-posed query. Examples of this phenomenon are provided in supplemental documentation. 

We evaluated Disharmony on virtual try-on method(StableVINTON\cite{stableviton}). We tested metrics such as SSIM and LPIPS on 100 pairs of edited and original images, which indicated that the background remained stable(refer to Table~\ref{tab:comparison_imagicinst}) suggesting that only the clothing in the images was edited. However, due to loose masking and the fact that our network was not specifically trained on this type of data, the network occasionally detected not only the edited clothing but also the upper part of the body. 

We also evaluated Disharmony using the drag-based method (DragDiffusion\cite{dragdiffusion}). We tested metrics such as SSIM and LPIPS on 10 pairs of edited and original images. From 10 different categories for validating the model, we randomly selected one image per category to generate the edited images. The SSIM and LPIPS metrics indicated that the background remained stable (refer to Table ~\ref{tab:comparison_imagicinst}) suggesting that only the dragged portions had been edited. However, due to our network not being trained on this specific diffusion model-based editing pipeline, and the absence of masks, which were replaced by general user edits and drag points, we propose further investigation into diffusion-based edits, including virtual try-on and drag-based edits, as future work.
A preliminary visual study of the various edits generated using the diffusion model, along with an evaluation using Disharmony, is provided in the supplemental documentation.
% \begin{table}[t!]
%     \centering
%     % \footnotesize
%     \begin{tabular}{lccc}
%         \hline
%         \textbf{Method} & \textbf{SSIM($\uparrow$)} & \textbf{PSNR} & \textbf{LPIPS($\downarrow$)} \\
%         \hline
%         Imagic & 0.343 & 11.60 & 0.601 \\
%         InstructPix2Pix & 0.456 & 14.15 & 0.543 \\
%         StableVITON & 0.831 & 15.95 & 0.188 \\
%         \hline
%     \end{tabular}
  
%     \caption{Comparison of Image Metrics for Imagic, InstructPix2Pix, ans StableVITON}
%     \label{tab:comparison_imagicinst}
% \end{table}

\begin{table}[t!]
    \centering
    \footnotesize
    \begin{tabular}{lcc}
        \hline
        \textbf{Method} & \textbf{SSIM($\uparrow$)} &  \textbf{LPIPS($\downarrow$)} \\
        \hline
        Imagic\cite{Imagic} & 0.343 & 0.601 \\
        InstructPix2Pix\cite{Instructpix2pix} & 0.456  & 0.543 \\
        StableVITON\cite{stableviton} & 0.831  & 0.188 \\
        DragDiffusion\cite{dragdiffusion} & 0.714 & 0.1818\\
        \hline
    \end{tabular}
  
    \caption{Comparison of Image Metrics for Imagic, InstructPix2Pix, StableVITON, and DragDiffusion}
    \label{tab:comparison_imagicinst}
\end{table}

\section{Conclusion}

In this paper, we have addressed the growing challenge of detecting AI-generated or edited images, particularly those involving the insertion and harmonization of objects within a scene. As generative AI continues to advance, the need for robust forensic methods capable of identifying subtle image alterations becomes increasingly critical. Our research demonstrates that training segmentation models on harmonization tasks can be viable solution for detection of edited regions, whether these edits are manually crafted or generated through deep learning models.

We introduced Disharmony, a model specifically designed to detect various forms of image edits, including those that are often overlooked by traditional forensic methods. By utilizing an aggregated dataset of harmonization techniques, our model outperforms existing forensic networks in identifying harmonized objects integrated into their backgrounds. Analysis on quantitative metrics (ROC AUC and mIoU) demonstrated consistently improved performance across images generated by all six evaluated harmonization methods compared to existing forensic models, indicating that Disharmony effectively detects the edited regions.

% The results of our comprehensive analysis highlight the effectiveness of our approach, providing a novel solution to the detection challenges posed by the sophisticated capabilities of modern generative AI. Our findings contribute to the broader field of digital forensics, offering a powerful tool for media authenticity and content verification in an era where image manipulation is increasingly accessible.

\begin{center}
\section*{Acknowledgements}
\end{center}
% This work was supported in part by NSF Awards 2243979 and  DOE AMAIS support.
This work was supported by the U.S. DOE Office of Science, Office of Advanced Scientific Computing Research, under award 76125: ``AMAIS - Advanced Memory to support Artificial Intelligence for Science''. The Pacific Northwest National Laboratory is operated by Battelle for the U.S. Department of Energy under contract DE-AC05-76RL01830, and by NSF Awards 2243979.

%%%%%%%%% REFERENCES
{\small
\bibliographystyle{ieee_fullname}
\bibliography{PaperForReview}
}

% \end{document}

%%%%%%%%% SUPPLEMENTARY 
\clearpage
\setcounter{page}{1}
\maketitlesupplementary
\appendix

\section{Results of Pretraining and Fine-Tuning}

\begin{figure*}[ht!]
\begin{center}
\includegraphics[width=0.9\linewidth]{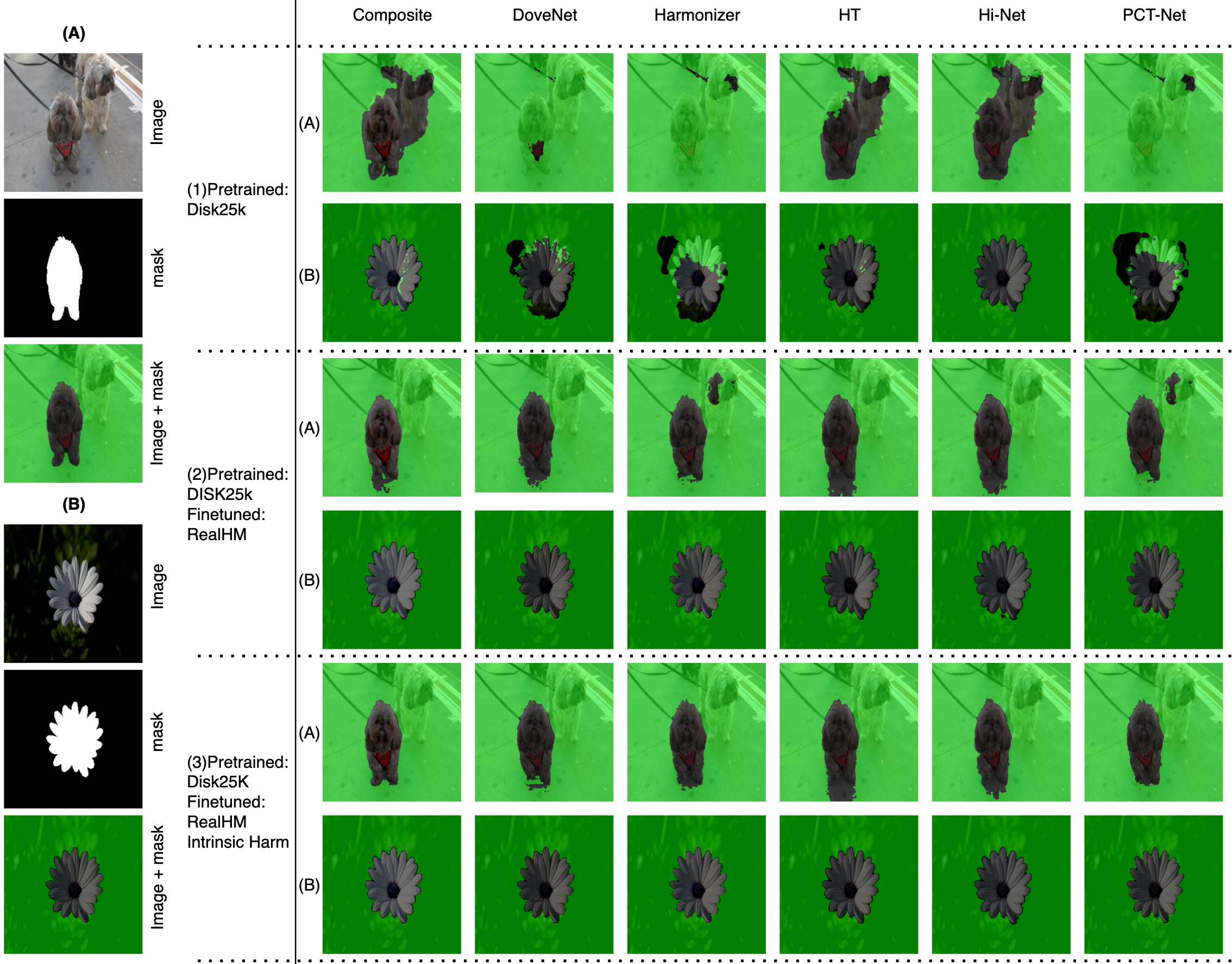}
\end{center}
\caption{Evaluation of the aggregation method. We tested three different scenarios: (1) pretraining MaskFormer\cite{MaskformerV1} with DISK25K\cite{Disk25kdata}, (2) pretraining with DISK25K followed by fine-tuning with the RealHM dataset\cite{RealHMDataset}, and (3) pretraining with DISK25K followed by fine-tuning with both the RealHM and IH dataset\cite{IntrinsicHarm}. Based on these evaluations, we concluded that the most effective approach is to pretrain with DISK25K and fine-tune with both the RealHM and IH dataset. }
\label{fig:SUPPTRAIN}
\end{figure*}

Fig.~\ref{fig:SUPPTRAIN}  illustrates the outputs generated by different training approaches: pretraining on DISK25K\cite{Disk25kdata} only, pretraining on DISK25K followed by fine-tuning with the RealHM dataset\cite{RealHMDataset}, and pretraining on DISK25K followed by fine-tuning with both the RealHM and IH\cite{IntrinsicHarm} datasets (the training approach used for our Disharmony).

When evaluating these models on a test set comprising various harmonization methods (DoveNet\cite{DoveNet2020}, Harmonizer\cite{Harmonizer}, HT\cite{IHT}, Hi-Net\cite{INR}, PCT-Net\cite{PCTNET}) and composite images, it is evident that the model pretrained solely on DISK25K struggled to accurately detect the harmonized objects. However, when fine-tuned with the RealHM dataset, the model's performance improved, producing more refined results, though some false detections persisted.

In contrast, our proposed training approach, which involves pretraining on DISK25K and fine-tuning with both the RealHM and IH datasets, effectively eliminated these false detections and produced even more refined segmentation masks. This demonstrates the validity of our training methodology, showing that combining datasets that are neural network-based, physics-based, and handcrafted light-aware can yield superior segmentation results.

% One might question why we did not train exclusively on the handcrafted dataset (RealHM) and the physics light-aware dataset (IH Dataset). However, two datasets are relatively small, containing only 216 and 60 images, respectively, which is significantly less than the 24,964 images in the DISK25K dataset. Thus, incorporating DISK25K was essential to achieve the comprehensive learning necessary for robust performance.

One might question why we did not train exclusively on the handcrafted (RealHM) and physics light-aware (IH) datasets. However, given their small sizes—216 and 60 images, respectively—compared to the 24,964 images in the DISK25K dataset, incorporating DISK25K was essential for achieving the comprehensive learning required for robust performance.

\section{Ground Truth Comparison}

To evaluate whether the forensic networks and Disharmony can correctly identify non-edited images, we tested them on the ground truth images from our generated test set. Our hypothesis was that if the images are not edited, the forensic networks should indicate the absence of edited regions. This experiment is significant, as previous studies have primarily focused on testing edited or generated image parts, leaving this aspect unexplored.

\begin{figure*}[ht]
\begin{center}
\includegraphics[width=0.85\linewidth]{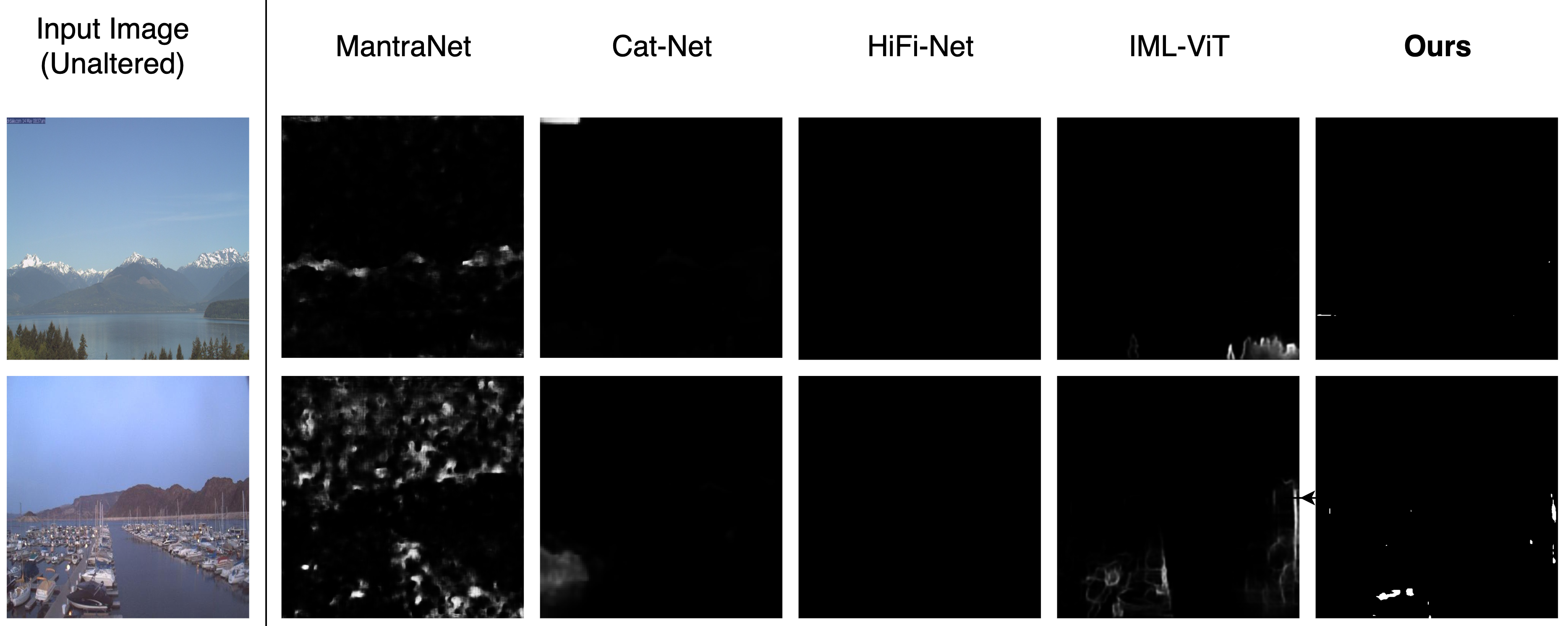}
\end{center}
\caption{The resulting masks produced by various forensic models(MantraNet\cite{Mantra-net}, Cat-Net\cite{CAT_net}, HiFi-Net\cite{HiFi-net}, IML-ViT\cite{IML-ViT}) and Disharmony, given an unaltered input image}
\label{fig:GroundTruth}
\end{figure*}

Fig. \ref{fig:GroundTruth} presents examples of the output from various networks, including our forensic network, when ground truth images were used as input. As shown, all networks except HiFi-Net\cite{HiFi-net} incorrectly identified edited regions in the non-edited images, resulting in false positive masks. Although our network also produced some false positives, it demonstrated greater stability across the ground truth images compared to the other networks. HiFi-Net, however, failed to detect any regions, which we will further discuss in Sec.~\ref{subsec:Qual_Comp} and Sec.~\ref{subsec:Quant_Comp} , highlighting why this outcome is problematic.

\section{Outputs from Text-Based Edits}

% As discussed in Sec.~\ref{Dis&Lim}, we generated test images for the Imagic TEDBench dataset, as illustrated in Fig. \ref{fig:SUPPLEMENTAL1}. At first glance, the edit instructions appear to be applied correctly, and the resulting images seem to align with the given prompts. However, upon closer examination, it becomes evident that the background, while maintaining the general context of the original image, has been significantly altered rather than just the intended part of the image. This leads to comprehensive changes to the original image, rather than the targeted edits intended. Additionally, we evaluated the Structural Similarity Index (SSIM), and Learned Perceptual Image Patch Similarity (LPIPS) for these images, as presented in Table \ref{tab:comparison_imagicinst}.

As mentioned in Sec.~\ref{Dis&Lim}, we generated test images using two text-based diffusion models, InstructPix2Pix\cite{Instructpix2pix} and Imagic\cite{Imagic}, on the TEDBench dataset. At first glance, the edit instructions appear to be applied correctly, with the resulting images seemingly aligning with the provided prompts. However, upon closer examination, it becomes clear that the background, while retaining the general context of the original image, has undergone significant alterations, rather than just the specific parts intended for editing. This results in comprehensive changes to the image as a whole, rather than the targeted edits that were intended. Additionally, we evaluated the SSIM\cite{SSIM} and LPIPS\cite{lpips} for these images, with the results presented in Tab.~\ref{tab:comparison_imagicinst}.

\section{Outputs from Virtual Try edits}

\begin{figure}[t!]
\begin{center}
\includegraphics[width=\linewidth]{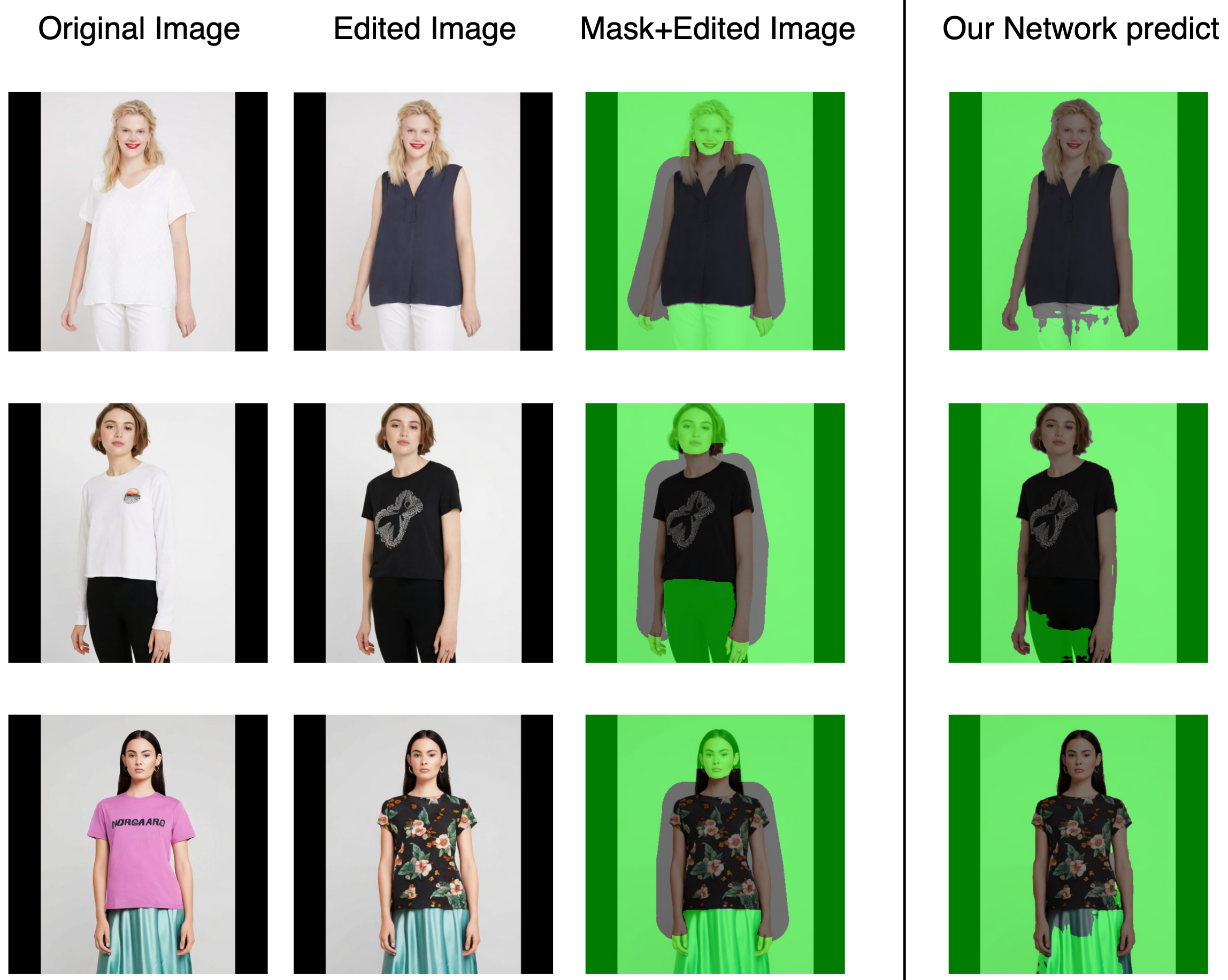}
\end{center}
\caption{The resulting segmentation outcomes for images that have been edited using StableViton\cite{stableviton}}
\label{fig:SUPPLEMENTAL2}
\end{figure}

As mentioned in Sec. \ref{Dis&Lim}, we conducted preliminary tests of our network on virtual try-on edits, acknowledging that further improvements are needed. The results are illustrated in Fig.~\ref{fig:SUPPLEMENTAL2}. In this experiment, StableVinton~\cite{stableviton} was used to generate an edited image where the same woman is shown wearing different clothes. We also present the mask used for training the model. Visually, the network appears to detect the regions corresponding to the clothing, albeit with some noise. Additionally, the network erroneously identifies the human face as an edited part of the image, which we attribute to the fact that our model was not specifically trained on this dataset. Despite these limitations, the results suggest potential for future research, particularly in refining the model's accuracy for virtual try-on edits.

\section{Outputs from Drag-Based edits}

\begin{figure}[t!]
\begin{center}
\includegraphics[width=\linewidth]{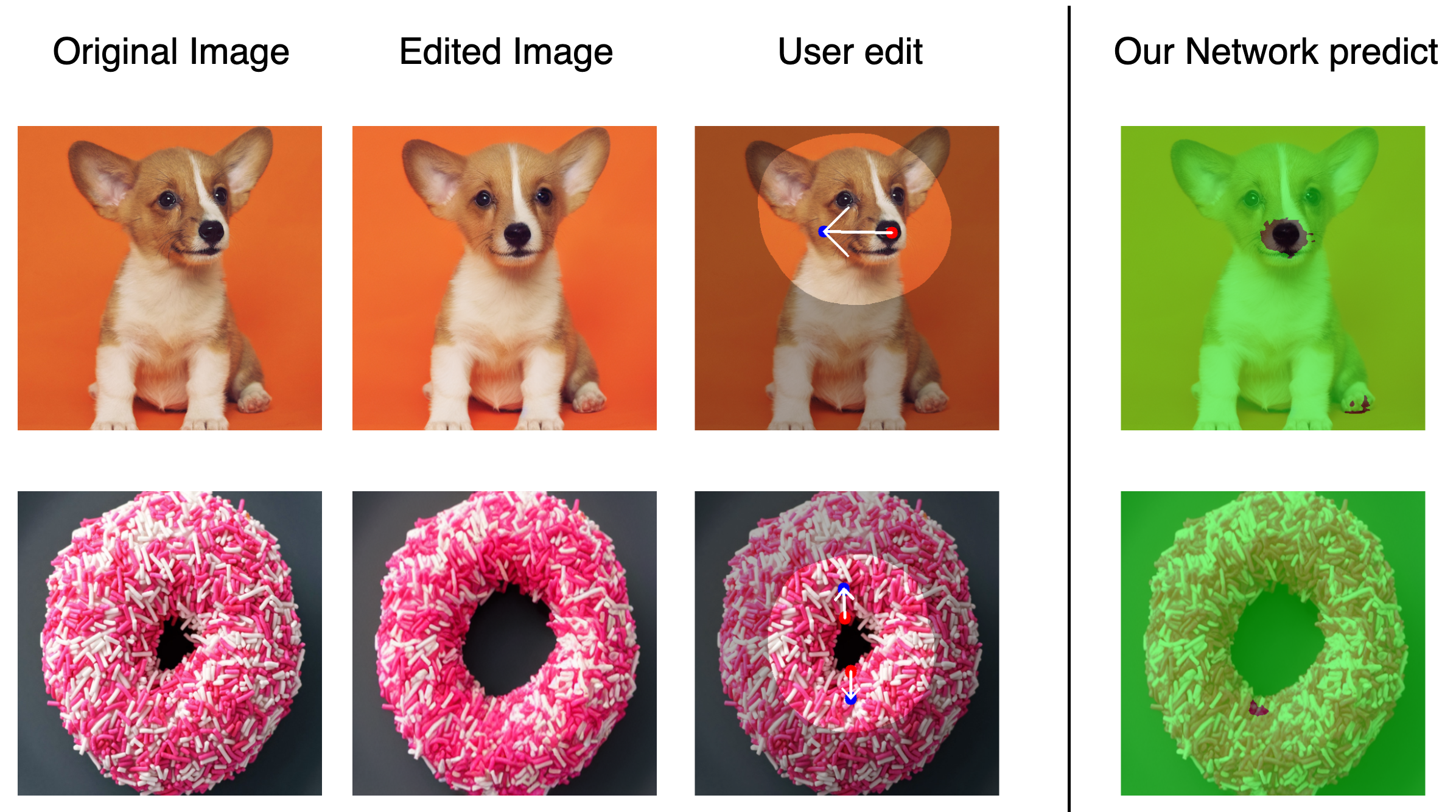}
\end{center}
\caption{The resulting segmentation outcomes for images that have been edited using DragDiffusion\cite{dragdiffusion}}
\label{fig:SUPPLEMENTAL3}
\end{figure}

As discussed in Sec. \ref{Dis&Lim}, we conducted preliminary tests of our network on drag-based edits, recognizing that further improvements are necessary. In this experiment, DragDiffusion\cite{dragdiffusion} was used to generate edited images based on user inputs. Upon running the network, we observed color shifting and blurring in some images, likely due to the diffusion process. However, our network was able to detect portions of the edited areas, although not consistently across the entire edited region. These findings suggest that there is potential for further exploration and refinement in this area. The results are illustrated in Fig.~\ref{fig:SUPPLEMENTAL3}.

\begin{figure}[ht]
\begin{center}
\includegraphics[width=\linewidth]{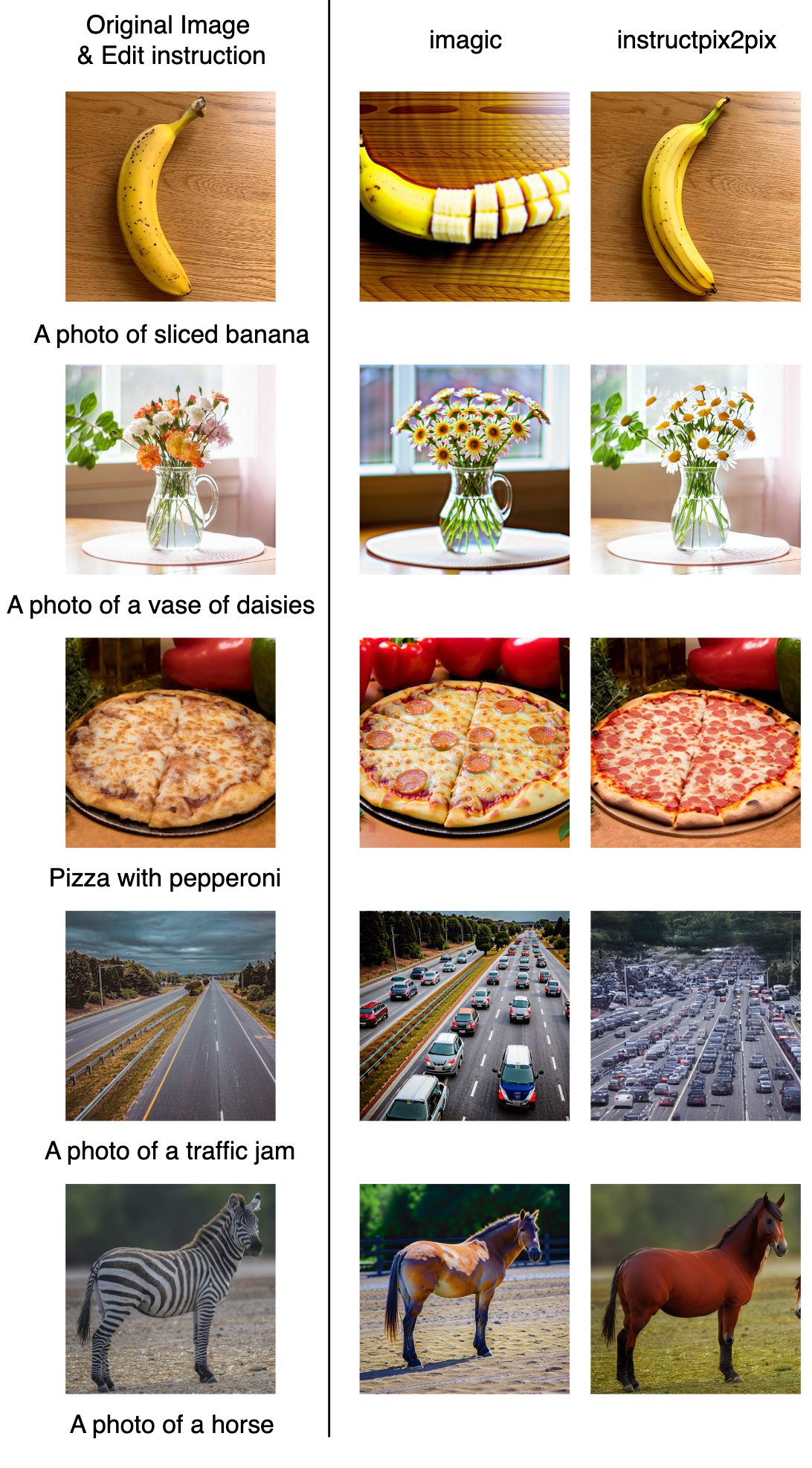}
\end{center}
\caption{The resulting images produced by text edits using InstructPix2Pix\cite{Instructpix2pix} and Imagic\cite{Imagic} on the TEDBench dataset. Note that the backgrounds change even when the text instructions do not specify modifications to the background.}
\label{fig:SUPPLEMENTAL1}
\end{figure}

\end{document}